# Off-line Programming and Simulation from CAD Drawings: Robot-Assisted Sheet Metal Bending

Pedro Neto
Department of Mechanical Engineering, CEMUC
University of Coimbra
POLO II, 3030-788, Coimbra, Portugal
pedro.neto@dem.uc.pt

*Abstract*—Increasingly, industrial robots are being used in production systems. This is because they are highly flexible machines and economically competitive with human labor. The problem is that they are difficult to program. Thus, manufacturing system designers are looking for more intuitive ways to program robots, especially using the CAD drawings of the production system they developed. This paper presents an industrial application of a novel CAD-based off-line robot programming (OLP) and simulation system in which the CAD package used for cell design is also used for OLP and robot simulation. Thus, OLP becomes more accessible to anyone with basic knowledge of CAD and robotics. The system was tested in a robot-assisted sheet metal bending cell. Experiments allowed identifying the pros and cons of the proposed solution.

*Keywords*—*CAD, off-line programming, simulation, robotics, sheet metal bending*

## I. INTRODUCTION

Software packages dedicated to OLP and robot simulation provide a set of features that make them capable to generate programs and simulate a given robotic task with some level of reliability. Berger *et al.* report the necessity to generate robot paths from CAD data of different free forming surfaces, using OLP software to parameterize and visualize the generated paths [1]. Bruccoleri *et al.* present an OLP approach for welding robots based on the integration of a software tool for robot simulation and automatic generation of robot programs [2]. Indeed, a major advantage of OLP and simulation has to do with its capacity to generate robot programs and predict system performance without taking the robot out of production during the programming phase.

Nevertheless, OLP software is an investment difficult to justify for most of small and medium sized enterprises (SMEs). Advantages of OLP are tempered by some limitations in existing software. In fact, most of OLP software is not intuitive to use, some are dedicated to a specific robot manufacturer and can only be applied (with reliability) in situations where the robot surrounding environment is known a priori and well modeled. In addition, robot calibration for OLP solutions continues to be a laborious task in which error is always present [3].

In recent years, CAD technology has become economically attractive and easy to work with. Millions of SMEs worldwide are using it to design and model their products. Over the years, some researchers have explored CAD technology trying to extend its capabilities to the robotics field. Today, it is possible to extract information from "raw" CAD drawings/files to generate robot programs. Diverse solutions have been proposed for the processes of spray painting and coating [4]. An important study in the field of CAD-based OLP presents a method to generate 3-D robot working paths for a robotic adhesive spray system for shoe outsoles and uppers [5]. An example of a novel process that benefits from robots and CAD versatility is the so-called incremental forming process of metal sheets [6]. A robotic CAD/CAM system that allows industrial robots to move along cutter location data without using any robot language is in [7]. A CAD-based and a digitizer-based OLP strategy are analyzed for the application to the deburring of gear transmission housings for aerospace applications [3]. Also, CAD-based OLP has been applied for the customization of processes in the textile industry [8].

As we have seen above, a variety of research has been conducted in the fields of CAD-, CAM- and VRML-based OLP and simulation. However, none of the studies/software so far has an effective and comprehensive solution for intuitive and cost-effective OLP directly interfacing with a with a commercial CAD package. In particular, this is what manufacturing system designers and robot integrators are looking for: the same CAD platform used for cell design, OLP and robot simulation.

### A. Proposed Approach

This study complements recent research in CAD-based OLP and simulation in which robot programs are generated from a CAD drawing running on a common 3-D CAD package, *Autodesk Inventor* [9-10]. In this paper, the same platform is applied to generate and simulate robot programs in a robot-assisted sheet metal bending cell. Fig. 1 resumes the architecture of the proposed system. Robot motion data are extracted from CAD drawings using an application programming interface (API) provided by *Autodesk*. These data are treated and used to generate robot working paths/programs for the application in study. The generated programs can be simulated in the same platform, path and arm motion simulation. Arm movement simulation is performed recurring to the *Robotics Toolbox for MATLAB*. In this case, *Autodesk Inventor* serves as graphical user interface (GUI). It was developed a software interface that allow users to manage the entire process.

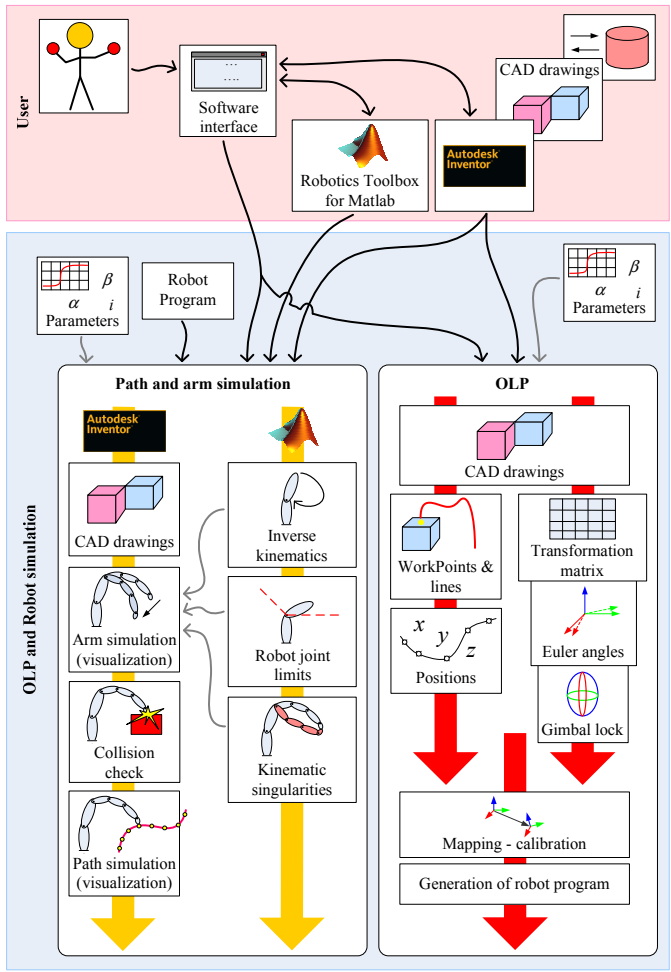

Fig. 1. Architecture of the CAD-based OLP and simulation system.

### B. Robot-Assisted Sheet Metal Bending

Increasingly, robots are being used in the metal bending industry performing monotonous and dangerous tasks (handling of sheet metal workpieces) that were previously performed by human labor. The introduction of robot manipulators into sheet metal bending cells allowed to improve quality, productivity, safety and reduce costs. However, the robotization of the process brings some difficulties such as the complexity and the large amount of time spent in robot programming task, possible collisions between the robot and the workpiece or other elements of the cell, and the need for skilled workers to operate and program robots. In this scenario, robot programs have to be semi-automatically and off-line generated. Teach-in robot programming becomes rapidly complicated, monotonous and time expensive with increasing part complexity. In addition, the specific characteristics of the bending process are liable to create problems such as the accumulation of errors throughout the process.

Over the last two decades, different solutions dedicated to automate the sheet metal bending process have been developed: process planning, path planning and definition of bending sequences [11]. Few studies refer to OLP and robot simulation considering the specificities of this process.

Generally, a robot-assisted sheet metal bending cell is composed by a CNC press brake with a set of punches and dies, a robot with a proper gripper/tool and a re-grasping/repositioning station, Fig. 2.

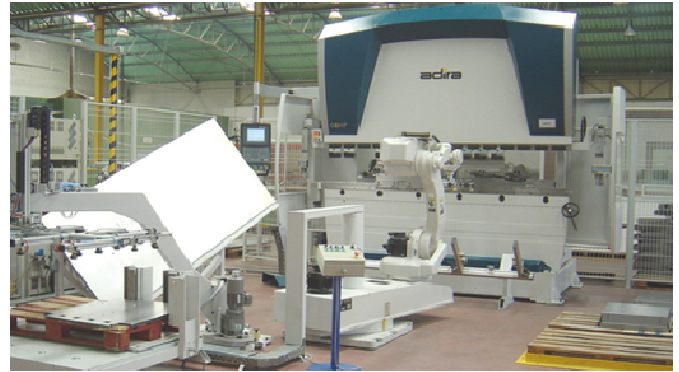

Fig. 2. A robot-assisted sheet metal bending cell (ADIRA SA).

There follows a general overview of the "common" stages of a robot-assisted sheet metal bending process:

*1) The robot picks up a flat sheet from the "input" pallet (sheet magazine table), or from an orientation station.*
*2) The bending operation begins, Fig. 3.*
*3) Re-grasping/repositioning operation, when the robot needs to change its grasp location between different bends.*
*4) The the bending process is complete, the robot places the finished workpieces in a desired pattern, "output" pallet.*

Fig. 3 shows details about the effective bending operation. Briefly the process works as follows:

*1) The workpiece firmly grasped is positioned on the die.*
*2) The punch approaches the workpiece.*
*3) The robot releases the workpiece.*
*4) The workpiece is formed due to the penetration of the punch into the die.*
*5) The robot grasps again the workpiece.*
*6) The punch goes up and the robot takes the workpiece.*

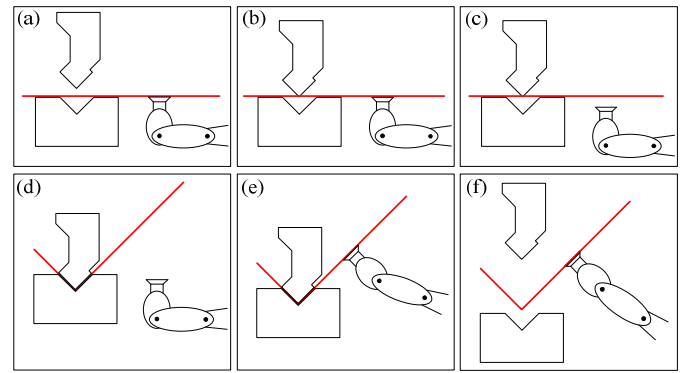

Fig. 3. The different stages of a bending operation.

In order to produce accurate bends and reduce scrap rates, the process design should predict some factors: bending sequences, collisions, selection of tools, material characteristics (spring-back), bendability, etc. Equally important is the quantification of the problems caused by spring-up of the workpieces, especially the buckling and deflection phenomenon. This assumes particular relevance in the bending of thin and long plates (high weight and low thickness). For this type of plates the robot must follow the movement of the workpiece during the forming process.

The process design occurs in subsequent iterations until to reach a feasible solution. The cost criterion is defined by each planner and can be based on different factors: handling time, number of tool changes, number of repositions, etc.

## II. CAD-BASED OLP

The developed software interface interacts with *Autodesk Inventor* using the *API* to extract data from CAD, generate robot programs and create actions within existing CAD drawings (simulation) [9]. It has a GUI with a set of end-user objects such as a CAD visualization window, a status alerting window and a set of menus where the user can easily define robot and process parameters, Fig. 4.

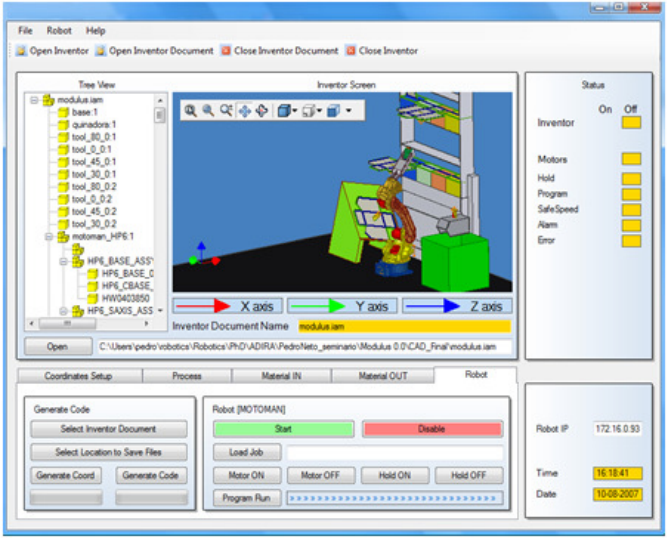

Fig. 4. Software interface: GUI.

### A. CAD Models and Planning

The general rules for creating the CAD models of the robotic cell in study are explained in [9]. These models have to be defined in order to allow extracting the positions and orientations that define the desired robot end-effector paths. Starting from the CAD *assembly model* of the robotic cell, the robot paths can be defined by introducing within the *assembly model* extra robot tool models (simplified models) representing the robot target points in a desired sequence. The *API* provides the transformation matrix of each tool model from which it is possible to extract the orientation and position of that tool. Fig. 5 shows a CAD drawing of a robotic bending cell from where a robot program is generated. Some specificity can be identified:

*1) The press brake is modeled as a "ladder" with one real bending line and many other virtual bending lines. This is because all robot motion relating to a manufacturing process is represented in only one assembly model. In consequence, if the press brake model has only one bending line (the real one) the drawing becomes very confusing, with several tool models in a situation of overlapping. The solution was to create a press brake with more bending lines, the virtual ones. In this way the user can represent the robot working paths by placing virtual tools along all the bending lines. The working sequence is defined by the name of the tool models (step_1, step_2 …). Since the distance between the bending lines is fixed and known, the software interface is prepared to interpret this situation and generates robot commands as if all the tool models were placed along the real bending line, Algorithm 1.*

*2) Robot poses are defined by the virtual tools within the CAD assembly model. For this specific application, considering for example the tool names step_2A and step_2B, the character "A" and "B" in the end of the name of the tools represent positioning (before the forming process) and releasing (after the forming process), respectively, Fig. 5 and Fig. 6.*

*3) In each different bending line the colored blocks represent the area of operation of each tooling set.*

It is during the construction of the CAD model of the cell that the user/designer defines a couple of issues associated with the planning of the process: the best grasping poses, ensures that no collisions occur, minimal deflection, etc. The bending sequences should also be planned with care, recurring to support software's or just to the know-how of an expert process designer. Fig. 7 shows a workflow of the process. The starting point is the CAD *assembly model* of the robotic cell in study. Then, some parameters inherent to the robot and process are defined by the user and robot programs are generated. These programs can be simulated and tested to check for abnormalities. If there are abnormalities and adjustments are necessary, the user can adjust the CAD model or the process and/or robot parameters. This process is repeated until to reach the desired performance for the robot program.

---

**Algorithm 1** Adjusting positions to the real bending line

**Input:** press brake height - *pbh*; distance between virtual bending lines - *dbl*; robot lower working area - *lwa*; robot upper working area - *uwa*; number of "ladder" steps - *nl*; number of press brake tools - *nwt*; robot end-effector position - *pos*;

**Output:** updated position - *u_pos*;

1: **Begin**
2:   **For** $n$ = 1 **To** *nl*
3:     **For** $i$ = 1 **To** *nwt*
4:       **If** $((pos(i).z - (n * dbl)) < (pbh + uwa))$
5:       **and** $((pos(i).z - (n * dbl)) > (pbh + lwa))$ **Then**
6:         $u\_pos(i).z \leftarrow pos(i).z - (n * dbl)$
7:       **End If**
8:     **End For**
9:   **End For**
10: **End**

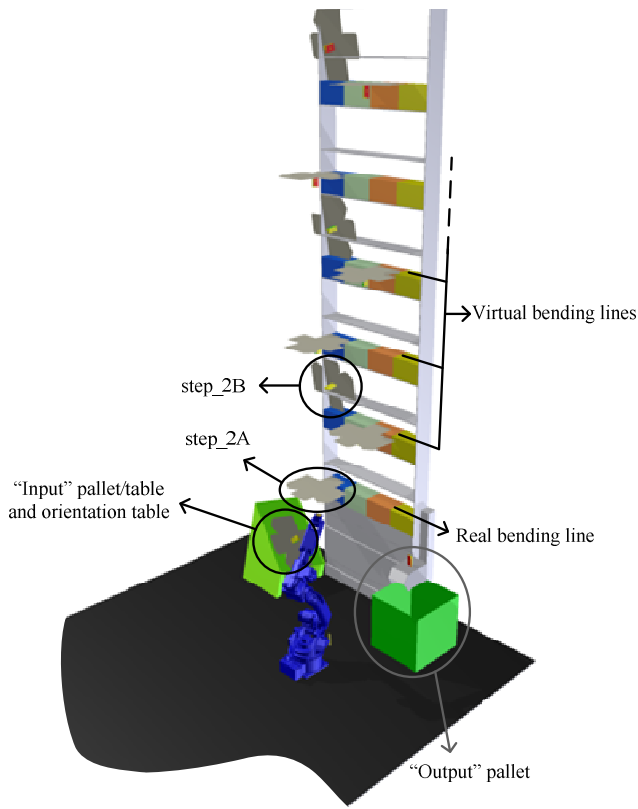

Fig. 5. CAD *assembly model* of a robot-assisted sheet metal bending cell.

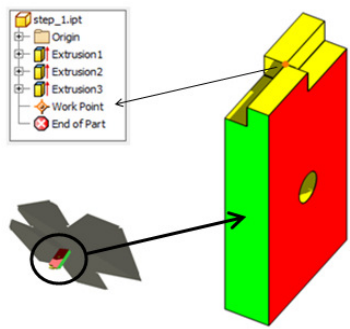

Fig. 6. Simplified tool model.

### B. Generation of Robot Programs

Robot programs are generated with base on an algorithm to generate robot commands from CAD drawings data. The proposed algorithm interprets that data and generates a program file. The algorithm and subsequent process of generation of a robot program is mainly divided into two distinct phases: definition and parameterization of robot positions/orientations, and, the definition of the body of the program that contains predominantly robot motion instructions. It is in this second phase that some specificity/customization to the process in study exists. For example, the generation of robot IO commands to communicate with the press brake.

It was established that the CAD *assembly model* should be simple to create. Thus, the user only needs to place two tool models to define a complete bend. A positioning and releasing pose before the forming process (step_*m*A) and a positioning and grasping pose after the forming process (step_*m*B). All needed related poses are automatically created by the algorithm having as input these two poses and the process/robot parameters introduced into the software interface, namely the approaching distances, Algorithm 2. For a simple robot movement the final robot pose is defined by a tool model named step_*g*F. The tool model named step_*last number* defines the pose for the first finished piece to be palletized Algorithm 2. Fig. 8 presents an example of the generated code for a *Motoman* robot.

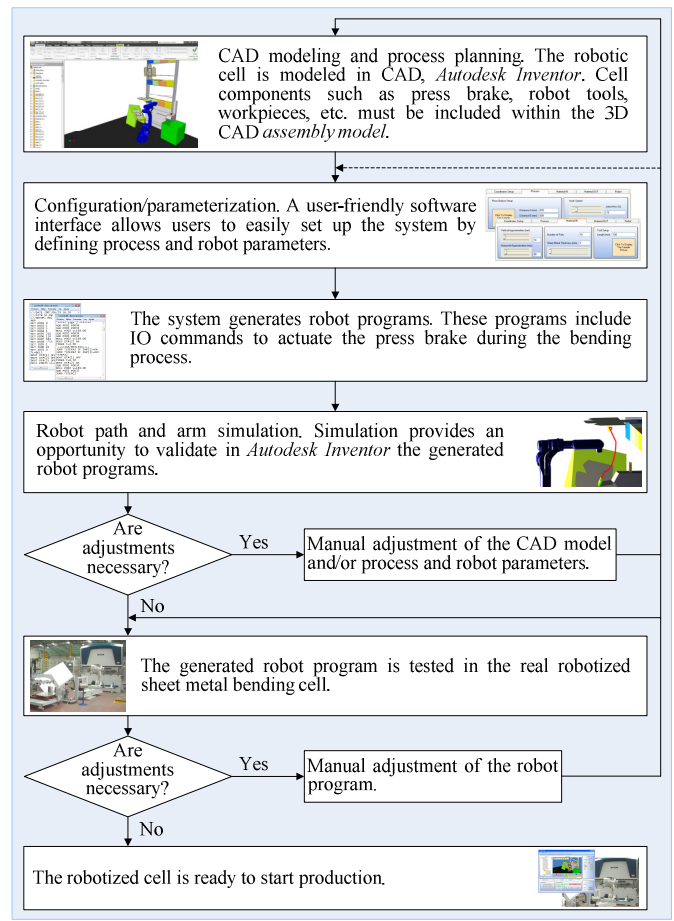

Fig. 7. Process workflow to generate a robot program from CAD.

```
'***** STEP 10 *******
ADD P0010 P0034
SUB P0010 P0035
MOVL P0010 V=169.00
SUB P0010 P0034
ADD P0010 P0035
MOVL P0010 V=169.00
DOUT OT#(2) ON
TIMER T=2.00
'+++ON/OFF++++
JUMP *STEPA10 IF IN#(1)=ON
JUMP *STEPB10 IF IN#(1)=OFF
*STEPA10
DOUT OT#(1) OFF
TIMER T=0.50
DOUT OT#(2) ON
```

Fig. 8. Generated robot code for a *Motoman* robot.

| Algorithm 2 Generation of robot programs |
|---|
| **Input:** CAD drawing of the robotic cell; |
| **Output:** robot program file (JBI file); |
| 1: **Begin** |
| 2:   **For Each** tool model **Do** |
| 3:     Get *WorkPoint* position |
| 4:     Get transformation matrix |
| 5:     Calculate frame correlations |
| 6:     Calculate Euler angles |
| 7:     **If** (tool name = step_1) **Then** |
| 8:       /* generate code for a picking up operation (increment) */ |
| 9:       Definition of robot pose |
| 10:       Definition of part of the body of the robot program |
| 11:     **Else If** (tool name = step_gF) **Then** |
| 12:       /* generate code for a simple robot movement */ |
| 13:       Definition of robot pose |
| 14:       Definition of part of the body of the robot program |
| 15:     **Else If** (tool name = step_*m*A) **Then** |
| 16:       /* generate code for positioning and releasing */ |
| 17:       Definition of robot pose |
| 18:       Definition of part of the body of the robot program |
| 19:       **If** (step_*m*B exists) **Then** |
| 20:         Get *WorkPoint* position (for step_*m*B) |
| 21:         Get transformation matrix (for step_*m*B) |
| 22:         Calculate frame correlations (for step_*m*B) |
| 23:         Calculate Euler angles (for step_*m*B) |
| 24:         /* generate code for positioning and grasping */ |
| 25:         Definition of robot poses |
| 26:         Definition of part of the body of the robot program |
| 27:       **End If** |
| 28:     **Else If** (tool name = step_last number) **Then** |
| 29:       /* generate code for shipment of finished products */ |
| 30:       Definition of initial robot pose |
| 31:       Definition of part of the body of the robot program |
| 32:     **End If** |
| 33:   **End For** |
| 34: **End** |

## III. EXPERIMENTS

Experiments involve OLP, simulation and tests with a real robot for the production of a specific workpiece. Results are analyzed and discussed. Since we do not have a press brake in laboratory, initial experiments were performed by using an industrial robot *Motoman HP6* equipped with an *NX100* controller and a second robot to handle the workpiece in study, making the workpiece to behave as when formed by a press brake. This second robot is a seven-axis robot *Motoman IA20*.

The workpiece used in the experiments is shown in Fig. 9. This is a common workpiece produced by the process in study and with multiple bends. Firstly, a robot program is generated from the CAD *assembly model* in Fig. 5.

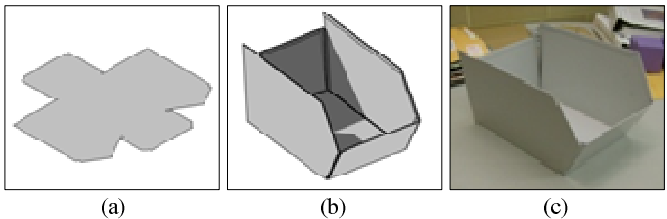

Fig. 9. Flat sheet model (a), final workpiece model (b), and real final workpiece (c).

### A. Simulation - Results

The robot program generated off-line from CAD is simulated in the same platform used to generate it, *Autodesk Inventor*. In this case, the proposed solution is able to interpret a given robot program and recurring to the *API* use the information contained into the program to produce actions within a CAD drawing, i.e., automate the process of drawing lines (virtual robot paths) and change drawing parameters (for example the robot joint angles). This last action allows to simulate robot arm movement. In this scenario the *Robotics Toolbox for MATLAB* is used to compute the inverse kinematics of robot models. Since most of companies do not have *MATLAB* software, in this moment they are using the system only for OLP and path simulation.

The simulated robot paths are displayed on the CAD *assembly model* of the cell, Fig. 10. In fact, the *Autodesk Inventor* provides the opportunity to visualizing models from any point in space, making zoom, etc. Major errors in robot paths can easily be visually identified. Robot arm motion simulation complements path simulation. Fig. 11 shows a sequence of frames captured from *Autodesk Inventor* while arm motion is simulated [12]. In this phase it can be analyzed the robot arm behavior and existing collisions.

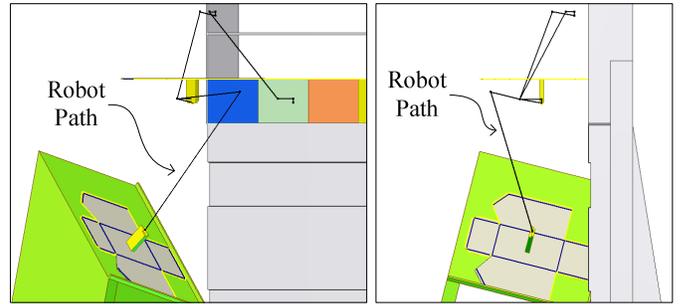

Fig. 10. Simulated robot paths from two different perpectives.

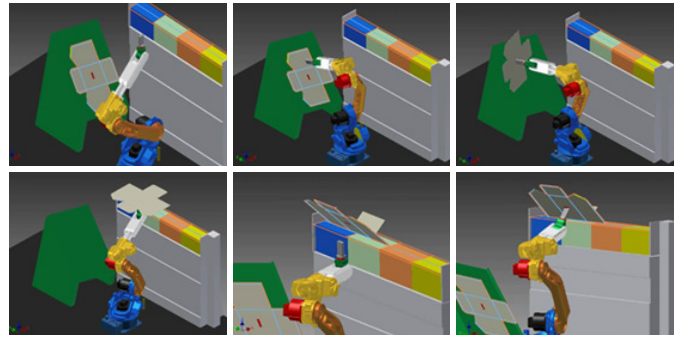

Fig. 11. Arm movement simulation.

### B. Real Robot - Results

Laboratorial experiments were performed using an industrial robot to "replace" the press brake, Fig. 12. These tests demonstrated that the algorithm to generate robot programs has all the necessary "main" functionalities for a correct definition of the bending process [12]. It was used a workpiece model in paperboard and a robot tool composed by a vacuum cup.

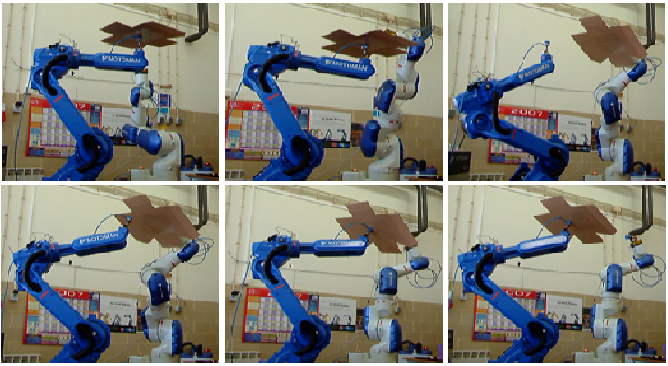

Fig. 12. A robot handling a workpiece prototype and another one reproducing the press behaviour into the workpiece.

*C. Discussion*

Experimental results proved that the generated robot programs are reliable to be used as a pre-program for a given robotic process, containing all the important steps and tasks that the robot has to perform. However, there are some negative situations/failures to highlight. These failures, which include collisions and part misalignment, occur due to part position uncertainty, errors in angle formation and in the length of the bent. This means that both the press brake and robot do not "know" the part's pose accurately in each instant of the process. Four different causes for this situation can be pointed out:

- Mechanical slop in the loading mechanism.
- Part pose information loss. This can happen when the gripper releases the part during the bending process.
- Slip during the grasping process and handling.
- Part buckling and deflection.

During the bending process the error is accumulative, after several bendings small errors may accumulate to critical values. In order to face this situation, the workpieces should be positioned in static reference systems (repositioning) to set the accumulated error to zero. Also, the CAD-based solution itself can present error from the calibration process or from situations where the models do not reproduce correctly the dimensions and geometry of the real scenario (this can be identified in the simulation). The above is also valid for commercial OLP software. This situation can be minimized by adding real-time sensory feedback to assist robots in their work [13]. Summarizing, the proposed CAD-based OLP system gives us a robot pre-program that needs to be tested and adjusted to the real robotic scenario. Anyway, this pre-program is a good help for robot programmers, allowing them to save time in the robot teach-in process.

## IV. CONCLUSIONS AND FUTURE WORK

This paper presented a novel CAD-based OLP and simulation system specifically tailored to generate programs for a robot operating on a sheet metal bending cell. Robot programs are generated from a CAD drawing running on a common 3-D CAD package, *Autodesk Inventor*. The advantage is twofold. First, robotic cell design and robot programming are embedded in the same interface and work through the same platform, *Autodesk Inventor*, without compatibility issues. Second, *Autodesk Inventor* is also used to simulate robot programs. Thus, product design, OLP and simulation are integrated seamlessly. This means that no advanced skills in robotics are needed to use such functionalities, only a minimum of robot specific knowledge and CAD are necessary. These are important issues to spread the utilization of robots and OLP solutions in SMEs.

Experiments showed that the proposed system is intuitive to use and has a short learning curve (modern CAD packages are intuitive to operate), allowing non-experts in robotics to create and simulate robot programs in just few minutes. Error and/or failures arise primarily from situations in which the CAD models do not reproduce correctly the real scenario, incorrect calibration or due to the mechanical slope of the workpieces during the bending process. Future work will be dedicated to improve the algorithm to generate code, making it more generalist, flexible and easier to tune. An ongoing work is the validation/customization of the system in/to industry.